\documentclass{article}

\usepackage{PRIMEarxiv}
\usepackage{amsmath} 
\usepackage[utf8]{inputenc} 
\usepackage[T1]{fontenc}    
\usepackage{hyperref}       
\usepackage{url}            
\usepackage{booktabs}       
\usepackage{amsfonts}       
\usepackage{nicefrac}       
\usepackage{microtype}      
\usepackage{lipsum}
\usepackage{graphicx}
\usepackage{caption}
\usepackage{multirow} 
\usepackage{amssymb}
\usepackage{tikz}

\usepackage{xcolor}
\usepackage{pgf}
\newcommand{\textred}[1]{{\color[RGB]{222,131,68}#1}} 
\newcommand{\textgreen}[1]{{\color[RGB]{126,171,85}#1}} 

\newcommand{\colornumber}[1]{%
  \ifdim#1 pt>0pt%
    \textcolor[RGB]{222,131,68}{#1}
  \else%
    \ifdim#1 pt<0pt%
      \textcolor[RGB]{126,171,85}{#1}
    \else%
      #1
    \fi%
  \fi%
}

\newcommand{\gradientnumber}[1]{
  \pgfmathsetmacro{\percentcolor}{max(min((#1 + 100)/200,1),0)} 
  \pgfmathsetmacro{\inversepercentcolor}{1 - \percentcolor} 
  \edef\newcolor{%
    \noexpand\definecolor{currentcolor}{rgb}%
      {\percentcolor, \inversepercentcolor, 0}
  }
  \newcolor 
  \textcolor{currentcolor}{#1}
}

\graphicspath{{media/}}     

\title{UniVG: Towards UNIfied-modal Video Generation 
}

\author{
  Ludan Ruan, Lei Tian, Chuanwei Huang, Xu Zhang, Xinyan Xiao \\
  Baidu Inc. \\
  Beijing, China\\
  \texttt{\{ruanludan, tianlei09\}@baidu.com} \\
  \texttt{huangcw21@gmail.com} \\
  \texttt{\{zhangxu44, xiaoxinyan\}@baidu.com} \\ 
}

\usepackage{color}
\newcommand\tianlei[1]{\textcolor{black}{#1}}

\newcommand\question[1]{\textcolor{black}{#1}}

\begin{document}
\maketitle
\vspace{-1cm}
\begin{figure}[ht]
    \centering
    \includegraphics[width=\linewidth]{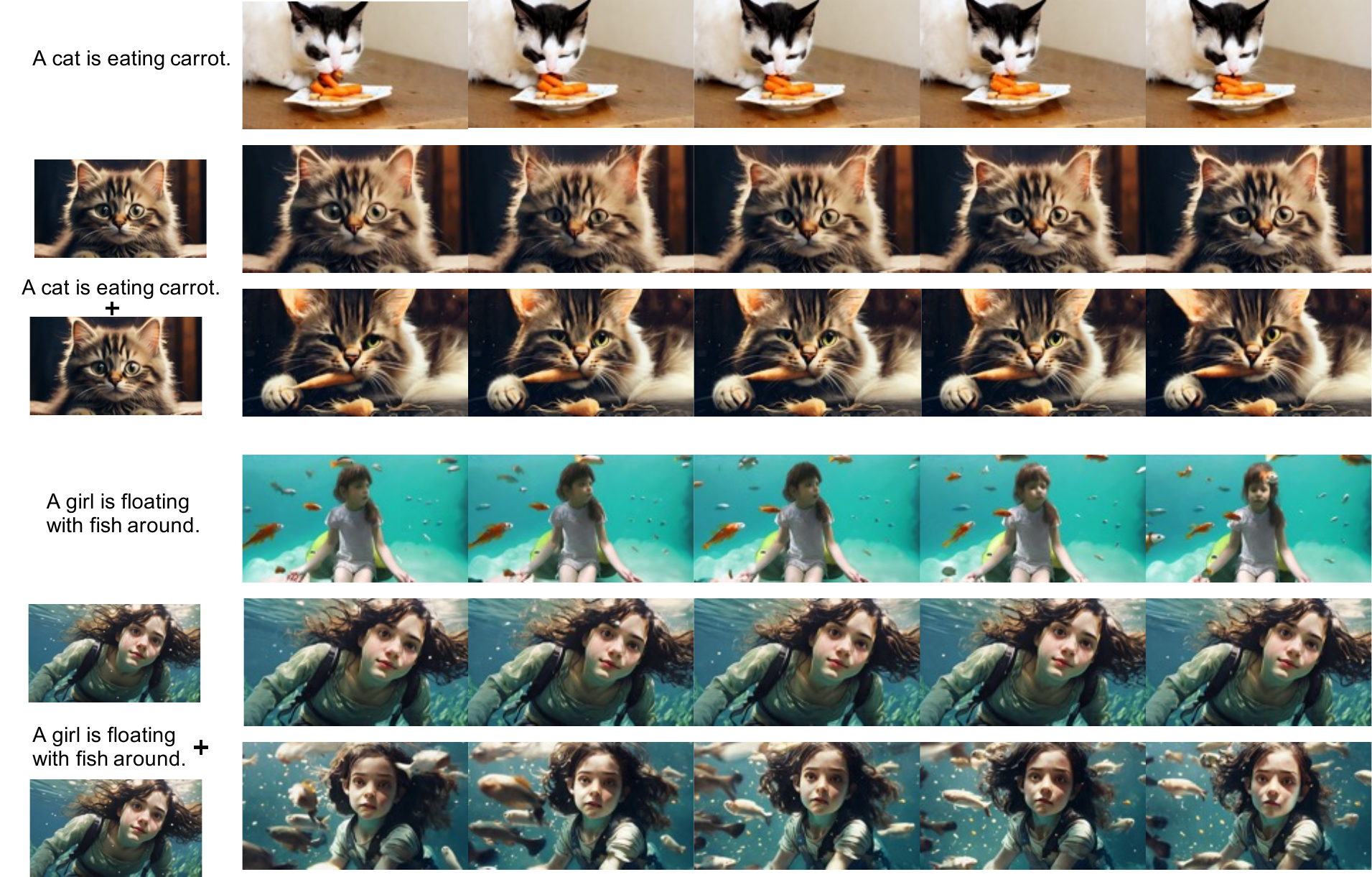}
    \caption{UniVG is a unified video generation framework that supports various video generation tasks, such as Text-to-Video, Image-to-Video, and Text\&Image-to-Video. Here displays two sets of examples. Row 1: Input text to generate semantically consistent videos; Row 2: Input image to produce pixel-aligned videos; Row 3: Combine the semantic of input text and image to create semantically aligned videos. All videos are shown on~\url{https://univg-baidu.github.io}. }

    \label{fig:teaser}
\end{figure}

\begin{abstract}
Diffusion based video generation has received extensive attention and achieved considerable success within both the academic and industrial communities. However, current efforts are mainly concentrated on single-objective or single-task video generation, such as generation driven by text, by image, or by a combination of text and image. This cannot fully meet the needs of real-world application scenarios, as users are likely to input images and text conditions in a flexible manner, either individually or in combination. To address this, we propose a \textbf{Uni}fied-modal \textbf{V}ideo \textbf{G}enearation system that is capable of handling multiple video generation tasks across text and image modalities. 
To this end, we revisit the various video generation tasks within our system from the perspective of generative freedom, and classify them into high-freedom and low-freedom video generation categories.
For high-freedom video generation, we employ Multi-condition Cross Attention to generate videos that align with the semantics of the input images or text. For low-freedom video generation, we introduce Biased Gaussian Noise to replace the pure random Gaussian Noise,  which helps to better preserve the content of the input conditions.
Our method achieves the lowest Fréchet Video Distance~(FVD) on the public academic benchmark MSR-VTT, surpasses the current open-source methods in human evaluations, and is on par with the current close-source method Gen2. For more samples, visit \url{https://univg-baidu.github.io}.

\end{abstract}

\section{Introduction}
In recent years, diffusion-based generative models~\cite{ho2020denoising,song2020score,song2020improved} have significant progress in image generation~\cite{dhariwal2021diffusion,nichol2021glide, ramesh2022hierarchical,saharia2022photorealistic,rombach2022high,wang2023videofactory} with applications rapidly expanding to video generation~\cite{singer2022make,ho2022imagen,girdhar2023emu,blattmann2023stable,guo2023animatediff}.
The majority of video generation models employ textual descriptions as conditional inputs~\cite{blattmann2023align,ge2023preserve,wang2023modelscope,hong2022cogvideo,singer2022make,ho2022imagen,he2022latent}. However, recent studies have begun to explore the use of image conditions to improve the detail of generated videos~\cite{chen2023videocrafter1} or for pixel-level controlling~\cite{li2023videogen,davids2023show1,blattmann2023stable,girdhar2023emu,zhang2023i2vgenxl,zeng2023make}. 
Additionally, to enhance the temporal smoothness and spatial resolution of the generated videos, current approaches often incorporate modules for frame interpolation and super-resolution~\cite{singer2022make,ho2022imagen}. 
However, existing works focus exclusively on single-objective or single-task video generation, where the input is limited to text~\cite{singer2022make,ho2022imagen,wang2023videofactory,hong2022cogvideo}, an image~\cite{blattmann2023stable}, or a combination of text and image~\cite{girdhar2023emu,zeng2023make}. This single-objective or single-task pipeline lacks the necessary flexibility to satisfy all user needs. In practice, users may not have the requisite text or image conditions for input, rendering the model unusable. 
Alternatively, the introduction of conflicting text-image pairs may lead to the generation of static videos or videos with abrupt transitions~(similar conclusion is proposed in~\cite{zeng2023make}).

In essence, all models used in video generation are conditional generative models that accept one or more conditions to produce a corresponding video. These conditions can be text, images, low-resolution videos, \tianlei{even control signals}.
In order to construct a versatile video generation system capable of handling multiple video generation tasks, we \tianlei{revisit existing methods and} categorize the relevant 
\tianlei{methods}
based on \textbf{generative freedom} rather than the task itself.
The concept of \textbf{generative freedom} that we propose corresponds to the 
\tianlei{range} 
of solution space for video generation models given certain conditions. In this paper, we categorize various video generation 
\tianlei{tasks}
as either high-freedom or low-freedom video generation.
\tianlei{
Specifically, high-freedom video generation is characterized by input conditions, i.e., text and image, that are weakly constrained at the semantic level, so that the generative model in this scenario has a larger solution space, providing a higher degree of freedom.
}
Conversely, low-freedom video generation typically involves strongly constrained conditions at  
\tianlei{the low-level information (i.e., pixel)}, such as in image animation and video super-resolution. These constraints limit the solution space available to the generative model, resulting in a lower degree of freedom.

\tianlei{In order to better match the characteristics of various video generation tasks,}
different strategies with varying degrees of generative freedom should be taken for video generation. 
For high-freedom video generation, the standard diffusion \textit{Generation Paradigm} is appropriate and has been extensively utilized in existing research~\tianlei{some refs should be provided @ludan}. Specifically, during training stage, the \tianlei{diffusion} model learns the added noise in the forward processing, and predicts the target distribution by reversing from a purely random Gaussian distribution during inference stage. Classifier guidance~\cite{dhariwal2021diffusion} and classifier free guidance~\cite{liu2023more} are employed to align the predicted distribution with the one specified by the input conditions.
For low-freedom video generation, the \textit{Editing Paradigm} is more suitable. Taking image editing~\cite{meng2022sdedit} as a case in point, a prevalent practice involves adding noise to the original image up to a certain level and then using text as the editing signal to steer the distribution toward the intended outcome. This approach, compared to generation from scratch, offers better retention of the original input's content. Video super-resolution has utilized a similar technique to that of image editing~\cite{zhang2023i2vgenxl}.
However, the \textit{Editing Paradigm} has a limitation in the form of a discrepancy between training \tianlei{stage} and inference \tianlei{one}. Specifically, the model is trained solely to approximate the target distribution without learning the transition from the conditional distribution to the target distribution.  
This discrepancy results in a trade-off-related issue, i.e., \textbf{the less noise that is introduced, the weaker the model's ability to edit, whereas the more noise that is added, the less capable the model is of preserving the input. }
In extreme cases, when the noise level approaches that of a completely random Gaussian distribution, editing \tianlei{paradigm} becomes analogous to generation \tianlei{one}, significantly diminishing the model's capability to preserve the content of the original input. 
How to reconcile the training and inference 
\tianlei{stages} of editing models to balance their editing capabilities while preserving the input is also a problem that needs to be addressed but has been overlooked in previous work.

In this paper, we propose a unified system \textbf{Uni}fied-modal \textbf{V}ideo \textbf{G}eneration~(i.e.\textbf{UniVG}), 
designed to support flexible video generation conditioned on the arbitrary combination of image and text. 
To achieve this, \question{we categorize all models within the system into two groups}: high-freedom video generation and low-freedom video generation. 
For high-freedom video generation, we present a base model that is capable of \tianlei{the requirements of handling} arbitrary combinations of text and image conditions. 
\question{We accomplish this by enhancing the original cross-attention module of the UNet architecture with a multi-condition cross-attention module.}
With regard to low-freedom video generation, we propose two 
\tianlei{corresponding} models that are individually tailored for image animation and video super-resolution task. These models utilize the editing paradigm, as opposed to the generation paradigm.  
\tianlei{To reconcile the differences between the training process based on generation paradigm and the inference process based on editing one}, 
\tianlei{
in this paper, we predict \textbf{B}iased \textbf{G}aussian \textbf{N}oise~(shorted as \textbf{BGN}) that is directed towards the target distribution, instead of standard Gaussian noise, by refining the objective function during training stage.
}

\tianlei{The proposed} UniVG system comprises a Base model, an Image Animation model and a Super Resolution model. 
The Base model 
\tianlei{is capable of handling} arbitrary combinations of text and image conditions and outputs \tianlei{a} video sequences of $24\times320\times576$ that are semantically aligned with the input conditions at 8 frames per second (fps). The Image Animation model that fine-tuned from the Base model with the additional condition of image \question{concatenation}, generates video frames of $24\times320\times576$ that are pixel-aligned with the input image. The Super Resolution model enhances the resolution of each frame to $720 \times 1280$ pixels.
Compared to previous works, 
\tianlei{Our UniVG demonstrates better tasks adaptability for video generation, i.e., handling various video generation tasks within an unified system,}
but also significantly improvements on the generation details and frame consistency.
Experiments have proven the effectiveness of our method. On objective metrics, our method significantly surpasses other existing methods, and in manual evaluations, our approach is on par with Gen2 and exceeds the other methods.

Our contributions can be summarized as follows:
\begin{enumerate}
\item We propose UniVG, the first video generation system that is capable of handling multiple video generation tasks, such as semantically aligned text/image-to-video generation, image animation.
\item We introduce Biased Gaussian Noise and confirm its effectiveness for low-freedom video generation tasks, such as image animation and super-resolution. 
\item Experiments demonstrate that our method surpasses existing text/image-to-video generation methods in terms of objective metrics and is on par with Gen2 in subjective evaluations.
\end{enumerate}

\section{Related Work}
\subsection{Text-to-Video Generation}
Early works on Text-to-Video generation utilized GANs~\cite{li2017video,pan2017create,yu2022digan}, VQ-VAEs~\cite{wu2021godiva,Mittal2017Sync}, auto-regressive models~\cite{wu2021godiva,hong2022cogvideo}, or transformer structure~\cite{ge2022tats}, but were limited by low resolution and suboptimal visual quality. Following the success of diffusion models in image generation~\cite{dhariwal2021diffusion,nichol2021glide,ramesh2022hierarchical,saharia2022photorealistic}, audio generation~\cite{Kong2021diffwave,chen2021WaveGrad,Popov2021GradTTS}, and other domains~\cite{gu2023NerfDiff,luo2023diffusion3d,ruan2023mmdiffusion}, VideoDiffusion~\cite{Ho2022videodiffusion} marked the first application of diffusion models in video generation. Subsequently, Make-A-Video~\cite{singer2022make} and ImagenVideo~\cite{ho2022imagen} expanded video generation into the open domain by extending the 2D U-Net from text-to-image generation to 3D U-Nets. Until then, researchers had been studying video modeling in the pixel space, which requires massive GPU memory consumption and high training costs. To address this issue, many researchers shifted their focus to conducting the diffusion process in the latent space instead of pixel space~\cite{rombach2022high,blattmann2023align,huang2022riemannian,ge2023preserve}, and to improving the sampling efficiency by learning-free sampling~\cite{song2020score,song2020denoising,liu2022pseudo,zhang2022fast} or learning-based sampling~\cite{watson2022learning,salimans2022progressive}. Additionally, some work has concentrated on reducing the training cost to that of a single video~\cite{Wu2022tuneavideo} or to no training cost at all~\cite{Khachatryan2023Text2VideoZero}.
\subsection{Image-to-Video Generation}
Generating video directly from text is a challenging task with high complexity. A natural thought is to use images as an intermediate bridge. Similar to Text-to-Video generation, early works on video prediction used non-diffusion methods~\cite{Rakhimov2021latentvideo,yan2021videogpt,tian2021mocoganhd}, which were often limited in low resolutions or specific domains. With the significant advancements in diffusion-based methods in Text-to-Video tasks, I2VGen-XL~\cite{zhang2023i2vgenxl} is, to our knowledge, the first to utilize diffusion for open-domain Image-to-Video generation. It replaces the textual CLIP features with image CLIP features within the text-to-video framework, achieving video generation semantically aligned with the input image. 
Similarly, SVD~\cite{blattmann2023stable} also fine-tunes from a text-to-video model to an image-to-video model but further concatenates the image's VAE features as a stronger controlling signal. Concurrently, videogen~\cite{li2023videogen}, VideoCrafter1~\cite{chen2023videocrafter1}, EMU Video~\cite{girdhar2023emu} and Make Pixels Dance~\cite{zeng2023make} remain their objective of text-to-video generation, but they introduce Text-to-Image synthesis as an intermediate step. The generated images are incorporated into the video generation framework either through concatenation or by CLIP features.

As can be inferred from the above, although text-to-video generation and image-to-video generation serve different applications, they share many similarities in their technical approaches. Therefore, this paper explores whether a single framework can unify these two objectives. The primary distinction of our UniVG from earlier works is that we differentiate various models included in video generation from the perspective of generative freedom rather than task. 

\section{Method}
\label{label:method}
This section presents our proposed \textbf{Uni}fied-modal \textbf{V}ideo \textbf{G}eneration~(i.e.~\textbf{UniVG}) for flexibly conditional video generation. Before diving into specific designs, we first briefly recap the preliminary knowledge of  diffusion models in Sec~\ref{label:method_diffusion}. We then illustrate the overview of the whole system UniVG in Sec~\ref{label:method_unimovg}, the \textbf{M}ulti-condition \textbf{C}ross \textbf{A}ttention~(i.e. \textbf{MCA}) used for high-freedom generation in Sec~\ref{label:method_multicond_cros_attn}, and the \textbf{B}iased \textbf{G}uassian \textbf{N}oise~(i.e. \textbf{BGN}) used for low-free generation in Sec~\ref{label:method_BGN}.
\begin{figure*}[t]
  \centering
  \includegraphics[width=\linewidth]{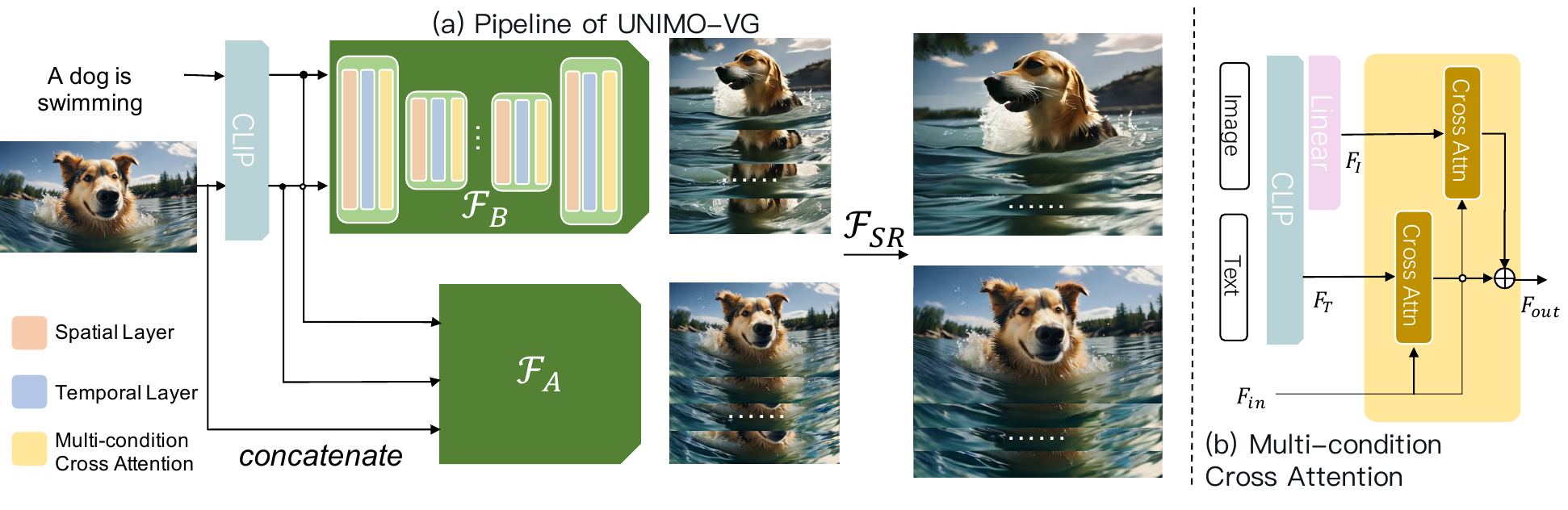}
  \caption{Overview of the proposed \textbf{UniVG} system. (a) displays the whole pipeline of UniVG, which includes the Base Model $\mathcal{F}_B$, the Animation model $\mathcal{F}_A$, and the Super Resolution model $\mathcal{F}_{SR}$. (b) illustrates the Multi-condition Cross Attention involved in $\mathcal{F}_B$ and $\mathcal{F}_A$. }
  \label{fig:unimo_vg}
\end{figure*}
\subsection{Preliminaries}
\label{label:method_diffusion}
Diffusion Models~\cite{ho2020denoising} are a class of generative models that are trained to generate the samples by iteratively denoising from Gaussian noise. 
During training, timestep $t (0 < t\leq N)$ determined noise is added at the original input $x$ to get noisy input $x_t = \sqrt{\overline{\alpha}_t} x_0 + \sqrt{1-\overline{\alpha}_t} \epsilon$~($\overline{\alpha}$ refers to noise schedule and $\epsilon$ refers to the noise that sampled from standard Gaussian distribution $\mathcal{N}(\mathbf{0},\mathbf{I})$), the model is trained to predict the added noise by either $\epsilon$-prediction~\cite{ho2020denoising} or v-prediction~\cite{salimans2022progressive}. 
During inference, samples are generated from pure noise $x_N \sim \mathcal{N}(\mathbf{0},\mathbf{I})$ by iteratively denoising. 
Furthermore, Conditional Diffusion Models~\cite{dhariwal2021diffusion,liu2023more} introduce extra conditional signals to bias the predicted distribution by $x_t = p_{\theta}(x_{t+1}) + w_c(p_{\theta}(x_{t+1},c)-p_{\theta}(x_{t+1}))$, where $\theta$ defines the diffusion model, $c$ defines input condition, and $w_c$ defines guidance scale of control intensity.
Another mainstream adopted diffusion models are Latent Diffusion Models~(LDM)~\cite{rombach2022high}, which consists of a Variational AutoEncoder~(VAE)~\cite{kingma2013auto} and a latent diffusion model that denoising in latent hidden space. 
This approach reduces the complexity of fitting distributions at high resolution. 
In this paper, each single model of UniVG is a Conditional Latent Diffusion Model. 
That is, the video $V$ consists of $F$ RGB frames is first compressed into latent space $X\in \mathbb{R}^{F \times C \times H \times W} $ with an image auto encoder, then input into UNet with one or multiple conditions~(text condition $T$, image condition $I$, and low resolution video $V^{lr}$).

\subsection{UniVG}
\label{label:method_unimovg}
As illustrated in Figure~\ref{fig:unimo_vg}-(a), our entire UniVG consists of three models: (1) A Base model $\mathcal{F}_B$ accepts any combination of text and image conditions for high-freedom video generation. (2) An Image Animation $\mathcal{F}_A$ model accepts text-image pairs to generated video aligned with input image in pixel level, and (3) a Super-resolution model $\mathcal{F}_{SR}$ for improving spatial resolution. 
Each model is a latent diffusion model with 3D UNet architecture composed of Spatial Layers, Temporal Layers, and Cross Attention Layers. Following previous works~\cite{singer2022make,blattmann2023stable}, the Spatial Layer consists of 2D Convolution layer and spatial transformers, while the Temporal Layer consists of 1D temporal Convolution layer and temporal transformers. 
The cross attention module is used to process semantic control signals, such as text and image feature.

(1) For the Base Model $\mathcal{F}_B$, we employ an image encoder that matches the text encoder of CLIP~\cite{radford2021learning} inspired by VideoCrafter1~\cite{chen2023videocrafter1}. To fully utilize the global semantics and local details of input image and text, we utilize all $K_I$ visual tokens $F_I=\{f^I_i\}_{i=0}^{K_I}$ and all $K_T$ text tokens $F_T=\{f^T_i\}_{i=0}^{K_T}$ from the last layer of CLIP ViT. 
To enable the model with the ability of processing more than one semantic features, we extend the original Cross Attention to Multi-condition Cross Attention and introduce its mechanism in Sec~\ref{label:method_multicond_cros_attn}. 
(2) In order to further generate videos that aligned with the input image at the pixel level, we train the Image Animation model $\mathcal{F}_A$ by finetuning $\mathcal{F}_B$ and concatenating the hidden space feature of the first frame as an additional condition. Because of the additional condition, the corresponding channel dimension of the initial convolution layer's kernel changes from $C$ to $2C$. We initialize the extra parameters to zero to preserve the performance of the original model.
Using either $\mathcal{F}_B$ or $\mathcal{F}_A$, we can obtain video frames of $24 \times 320 \times 576$. 
(3) To upscale the clarity of the generated videos, we further finetune a Super-Resolution model $\mathcal{F}_{SR}$ from $\mathcal{F}_{B}$. 
Since super-resolution tasks have no image condition, the multi-condition cross attention module reverts to a regular cross-attention module that only accepts the text condition.
During training, $\mathcal{F}_{SR}$ accepts videos of low resolution $V^{lr}$, which are obtained by destroying high-definition videos through RamdomBlur, RandomResize, JPEG Compression and so on. 
As we classify the tasks corresponding to $\mathcal{F}_{A}$, and $\mathcal{F}_{SR}$ as low-freedom generation, we present the Biased forward and backward processes from conditional distribution to target distribution by adjusting the standard Gaussian Noise to Biased Gaussian Noise~(\textbf{BGN} that is introduced in Sec~\ref{label:method_BGN}).

\subsection{Multi-condition Cross Attention}
\label{label:method_multicond_cros_attn}
Since our base model $\mathcal{F}_B$ and Image Animation model $\mathcal{F}_A$ accept text and image CLIP features, we use Multi-condition Cross Attention instead of the standard Cross Attention. This module's architecture mainly follows VideoCrafter~\cite{chen2023videocrafter1}, which computes $F_{\text{out}}$ by:
\begin{align*}
    F_{out} = \text{Softmax}\left(\frac{Q_{in} K_{T}^\intercal}{\sqrt{d}}\right) & \cdot V_{T} + \text{Softmax}\left(\frac{Q_{in}K_{I}^\intercal}{\sqrt{d}}\right) \cdot V_{I} \\
    Q_{in} = W_Q \cdot F_{in},\ K_{T} = W_{K_T} \cdot F_{T},\ V_{T} &= W_{V_T} \cdot F_{T},\ K_{I} = W_{K_I} \cdot F_{I},\ V_{I} = W_{V_I} \cdot F_{I}
\end{align*}
where $d_k$ is the dimensionality of the key/query vectors and $Q_{\text{in}}$ is shared between $F_I$ and $F_T$. The weights $W_{K_I}$ and $W_{V_I}$ are initialized from $W_{K_T}$ and $W_{V_T}$, respectively. 
Unlike VideoCrafter1 that treats image as an additional input enhancement, we regard the image as an equally significant control signal along with the text. This is achieved by applying a certain proportion of image dropout throughout the training process.
By extension, MCA can accommodate more than two conditions by increasing the number of cross-attention units, without the necessity for retraining~(e.g. stronger text features). This flexibility greatly reduces the cost of extending the model's training to handle new conditions.
\begin{figure*}[t]
  \centering
  \includegraphics[width=\linewidth]{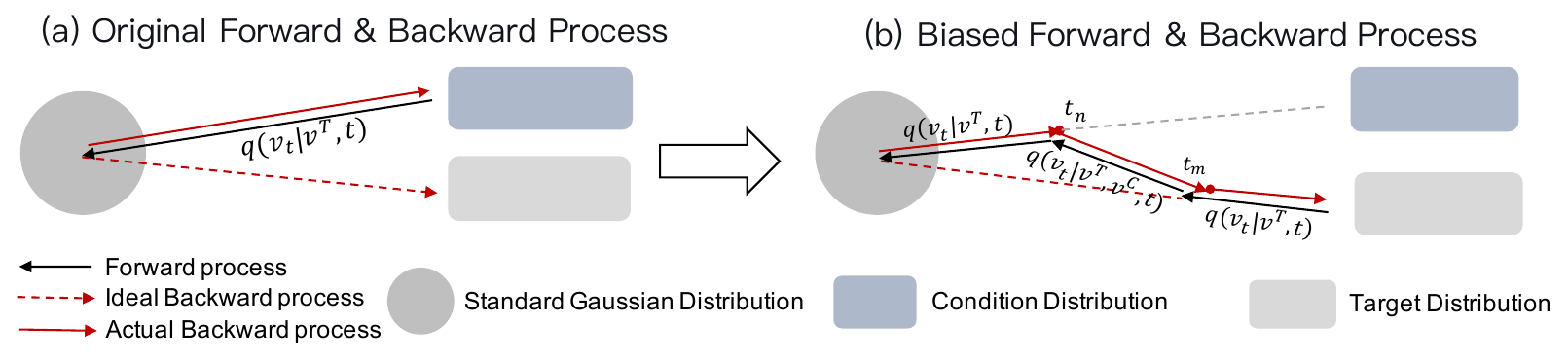}
  \caption{ The forward \& backward diffusion process with Random Gaussian Noise and Biased Gaussian Noise.}
  \label{fig:BGN}
\end{figure*}
\subsection{Biased Gaussian Noise}

\label{label:method_BGN}
Our proposed Biased Gaussian Noise is used to transfer condition distribution to target distribution for low-freedom video generation. 
As illustrated in Figure~\ref{fig:BGN}-(a), the standard forward diffusion process transitions from the target distribution $v^T$ to the standard Gaussian distribution $\epsilon$ via $v^T_t = \sqrt{\overline{\alpha}_t }v^T + \sqrt{1-\overline{\alpha}_t} \epsilon$. 
However, typically in the backward process, these are the only two distributions involved. This can result in suboptimal editing outcomes when the samples are introduced from a condition distribution $v^C$ during inference.  
To account for the condition distribution in both forward and backward processes, we segment the original diffusion into three parts, as illustrated in Figure~\ref{fig:BGN}-(b). 
For timestep between $0$ to $t_m$, $v_t$ is calculated by the target sample with $q(v_t|v^T,t) = \sqrt{\overline{\alpha}_t} v^T_0 + \sqrt{1-\overline{\alpha}_t} \epsilon (0 \leq t < t_m)$ that followed the original forward process. 
For timestep between $t_n$ to $N$, $v_t$ is calculated by the condition sample with $q(v_t|v^C,t) = \sqrt{\overline{\alpha}_t} v^C + \sqrt{1-\overline {\alpha}_t} \epsilon (t_n \leq t < N)$. 
The core problem is how to design $q(v_t|v^C, v^T, t)$  that can smoothly transition from $v_{t_m}$ to $v_{t_n}$. 
To preserve the original diffusion schedule, we introduce a variable for the noise $\epsilon$, denoted as $\epsilon'$.
Assume that for timesteps between $t_m$ and $t_n$, we have $q(v_t|v^C, v^T, t) = \sqrt{\overline{\alpha}_t}v^T + \sqrt{1-\overline{\alpha}_t}\epsilon'$, which meets the conditions $v_{t_m} = \sqrt{\overline{\alpha}_{t_m}}v^T_0 + \sqrt{1-\overline{\alpha}_{t_m}}\epsilon$ and $v_{t_n} = \sqrt{\overline{\alpha}_{t_n}}v^C_0 + \sqrt{1-\overline{\alpha}_{t_n}}\epsilon$.
Thus, the corresponding $\epsilon '$ should satisfy the following formulas at timestep $t_m$ and $t_n$.
\begin{align*}
    \epsilon_{t_m} ' = \epsilon, \ \ \epsilon_{t_n} ' = \epsilon + \frac{\sqrt{\overline {\alpha}_{t_n}} }{\sqrt{1-\overline {\alpha}_{t_n}} } \times \left( v^C - v^T \right)
\end{align*}
In theory, there are an infinite number of solutions to $ \epsilon '$. In this paper, we simply define $ \epsilon '$ as a linear transformation following 
\begin{align*}
    \epsilon_{t} ' = \epsilon + \frac{\sqrt{\overline {\alpha}_{t}} }{\sqrt{1-\overline {\alpha}_{t}} } \times \frac{t-t_m}{t_n-t_m} \times \left( v^C - v^T \right), \ (t_m \leq t <t_n)
\end{align*}
The $\epsilon'$ is sampled from a Biased Gaussian distribution, with its mean value shifted by a weighted combination of $v^C$ and $v^T$. This bias is crucial to bridging the diffusion process from the condition distribution to the target distribution. Alternative solutions for $\epsilon'$ will be explored in our future work.

\section{Experiments}
\subsection{Implementation Details}
\paragraph{Dataset} 
Our training datasets include publicly available academic datasets such as WebVid-10M~\cite{Bain21} and LAION-COCO~\cite{laioncoco}, along with self-collected data. 
WebVid-10M is a large and diverse text-video dataset containing approximately 10 million open-domain videos with a resolution of $336 \times 596$ pixels. LAION-COCO is a substantial text-image dataset comprising 600 million high-quality images, filtered from LAION-2B and scored using the Aesthetic and Semantic Estimate (ASE).
To further enhance the quality of the generated videos and to address the issue of watermarks present in WebVid-10M, we continue training on our own curated datasets of videos and images, which contain high-quality visual content. We prepare the self-collected videos by first proportionally compressing them to 720p resolution along their shorter edge and then segmenting them into 10-second clips. This process yielded 5 million high-quality text-video pairs. Additionally, our self-curated image dataset includes 1.3 million high-quality text-image pairs, with a focus on artistic styles.

\paragraph{Training}
Our  $\mathcal{F}_B$ is trained with an image:video:video frame ratio of 1:1:1, where the training video frames were sampled with equal probability from 8 to 24 frames. 
We set the text dropout to 0.5 and the image dropout to 0.1. 
In addition, we utilize offset noise~\cite{offsetnoise} with a strength of 0.1 and zero terminal Signal-to-Noise Ratio (SNR)~\cite{girdhar2023emu}. 
Offset noise has been proven helpful to be helpful in generating extremely dark or bright images.
Zero terminal-SNR has been shown to be beneficial for generating high-quality and high-resolution visual content by adding noise to pure Gaussian noise following a rescaled schedule.
Both techniques have proven useful in our experiments. 
Subsequently, we continue finetuning $\mathcal{F}_B$ to obtain $\mathcal{F}_A$ and $\mathcal{F}_{SR}$, using Biased Gaussian Noise (BGN) on our self-curated video dataset only.
For $\mathcal{F}_A$, we set the text dropout to 0.1 and the image dropout to 0.1, the BGN is experimentally set during timesteps $t_m=600$ to $t_n=990$ since the earlier steps determine the content~\cite{meng2022sdedit}.
For $\mathcal{F}_{SR}$, the text dropout is set to 0.1, and the BGN is applied during timesteps $t_m=0$ to $t_n=700$ since the later steps deciding the details~\cite{meng2022sdedit}.
We incorporate $\epsilon-$prediction~\cite{ho2020denoising} for $\mathcal{F}_B$ and  $\mathcal{F}_A$, $v-$prediction for $\mathcal{F}_{SR}$. The learning rate of all models is fixed at $1 \times 10^{-5}$.
We use DPM Solver~\cite{lu2022dpmsolver} for accelerating sampling: 50 steps for $\mathcal{F}_B$ and $\mathcal{F}_A$, and 7 steps for $\mathcal{F}_{SR}$ since we set initial weight to 0.7.

\paragraph{Evaluation}
We use both objective metrics and human evaluation as the assessment criteria for UniVG.
In terms of objective metrics, we follow the previous work~\cite{li2023videogen, zeng2023make} to use the test set of MSR-VTT~\cite{xu2016msr} as the standard benchmark. This test set comprises $2,990$ test videos, each corresponding to $20$ prompts, totaling $59,800$ prompts. For efficiency in our ablation study, we randomly selected one prompt for each test video, ultimately obtaining $2,990$ prompts as the evaluation set. We calculate the CLIPSIM~\cite{wu2021godiva} between the generated videos and the corresponding text, and FVD~\cite{feichtenhofer2021fvd} between the generated videos and the original videos as comparative metrics.
Since some studies~\cite{girdhar2023emu} have pointed out that objective metrics may not always align with human perception, we primarily employ human evaluation. 
Specifically, we adopt the categorization of video generation metrics from EMU video~\cite{girdhar2023emu}, which includes \textbf{V}isual \textbf{Q}uality~(including Visual Quality consists of pixel sharpness and recognizable objects/scenes), \textbf{M}otion \textbf{Q}uality~(including frame consistency, motion smoothness and amount of motion), \textbf{T}ext \textbf{F}aithfulness~(Includes text-spatial alignment and text-temporal alignment). 
Since UniVG supports conditional generation for any combination of text and image, we further introduce \textbf{I}mage \textbf{F}aithfulness~(Includes text-spatial alignment and text-temporal alignment) to measure the alignment performance of generated videos with given images. Evaluators also provide their \textbf{O}verall \textbf{L}ikeness of the two videos being compared, serving as a complement to the aforementioned sub-indicators.
The prompts used for human evaluation were collected from the webpages of previous work~\cite{singer2022make,ho2022imagen,li2023videogen,zeng2023make,ge2023preserve}, totaling $131$ in number.  
To simplify the annotation process, annotators need only select G (our method is better), S (equally good), or B (other methods are better) for each indicator. To ensure fairness, the videos being compared are randomized during the annotation process. Our six annotators provide a total of $6 \times 131$ (prompts) $\times 10$ (sub-metrics) = $7,860$ evaluation results.

\subsection{Comparison with SOTA}
\noindent 
\begin{minipage}[c]{0.5\textwidth} 
\centering
 \captionof{table}{Zero-shot performance comparison on MSR-VTT.
T refers that the input condition contains text and I refers to image. UniVG-HG refers to high-freedom generation within our UniVG, UniVG-LG refers to low-freedom generation within our UniVG.
Best in \textbf{bold}}
  \begin{tabular}{l|ccc}
    \hline
    Method & Input & CLIPSIM$\uparrow$ & FVD$\downarrow$  \\ \hline
    CogVideo(En)~\cite{hong2022cogvideo} & T & 0.2631 & 1294 \\ 
    MagicVideo~\cite{zhou2022magicvideo} & T & -  & 1290 \\ 
    LVDM~\cite{he2022latent} & T & 0.2381 & 742 \\
    Video-LDM~\cite{blattmann2023align} & T & 0.2929 & - \\
    InternVid~\cite{wang2023internvid} & T & 0.2951 & - \\
    Modelscope~\cite{wang2023modelscope} & T & 0.2939 & 550 \\
    Make-a-Video~\cite{singer2022make} & T & 0.3049 & - \\
    Latent-Shift~\cite{an2023latent} & T & 0.2773 & - \\
    VideoFactory~\cite{wang2023videofactory} & T & 0.3005 & - \\ \hline
    PixelDance~\cite{zeng2023make} & T+I & 0.3125 & 381 \\
    Videogen~\cite{li2023videogen} & T+I & 0.3127 & - \\ \hline
    UniVG-HG & T & 0.3014 & 336 \\
    UniVG-HG & T+I & 0.3136 & 331 \\
    UniVG-LG & T+I & \textbf{0.3140} & \textbf{291} \\
    \hline
  \end{tabular} \\

 \label{tab1_auto_metric}
\end{minipage}%
\hspace{2mm} 
\begin{minipage}[c]{0.48\textwidth} 
\centering
\includegraphics[width=0.9\textwidth,keepaspectratio]{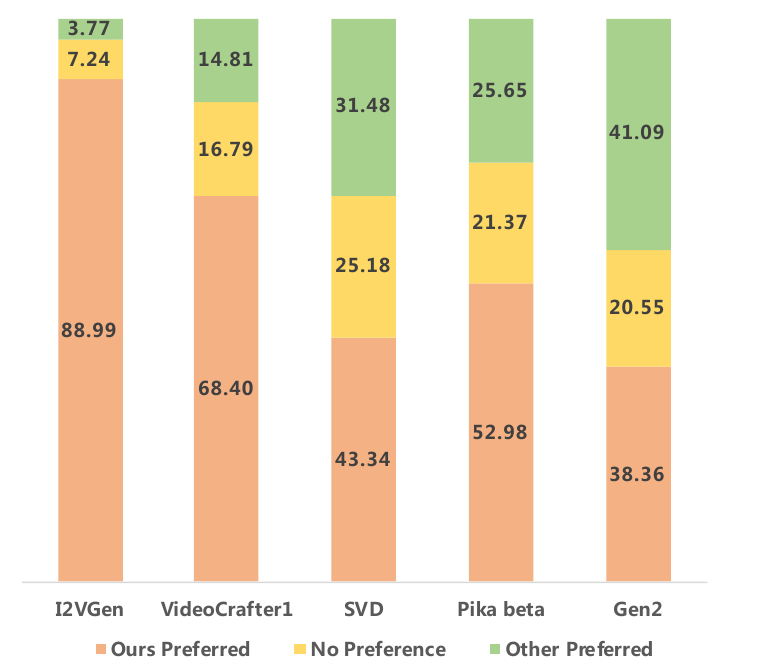}
\captionof{figure}{Percentage(\%) of Overall Preference of UniVG-LG generated videos compared with other SOTA methods.}
\label{fig1:prefernce}
\end{minipage}

\paragraph{Automatic Metrics}
Due to the previous work involving both plain text-to-video and image-to-video generations, we adopt aligned settings to conduct a fair comparison with them. For text-to-video generation, we use only text as the input condition to generate videos($\mathcal{F}_B$+ $\mathcal{F}_{SR}$). For image-to-video generation, we start by creating images from prompts using SDXL1.0+refiner~\cite{podell2023sdxl} and then proceed with both high-free generation~($\mathcal{F}_B$+ $\mathcal{F}_{SR}$) and low-free generation~($\mathcal{F}_A$+ $\mathcal{F}_{SR}$) using UniVG for the combination of text and images. Since the i3d model~\cite{carreira2017i3d} used for testing FVD can only accept 16 frames, we random sample 16 frames from our generated 24 frame and the test videos in MSR-VTT.
The results are shown in Table~\ref{tab1_auto_metric}.
Whether utilizing only text as the input condition or using both text and image together, our method generates videos that outperform those created by other methods under the same settings. 
Even when using only text as the condition, the videos generated by our method surpass in the FVD metric those generated by others that use both text and image. This demonstrates the capability of UniVG to concurrently process text and image conditions and generate high-quality videos with both good visual content and text alignment.

\begin{table}[h]
\caption{The winning rate~(\%) of UniVG-LG compared to other methods in human evaluations across 10 sub-dimensions(The abbreviations include VQ: Visual Quality, MQ: Motion Quality, TF: Text Faithfulness, IF:Image Faithfulness, OL: Overall Likeness, PS: Pixel Sharpness, RO/S: Recognizable Objects/Scenes, FC: Frame Consistency, Motion Smoothness, AM: Amount of Motion, TSA: Text-Spatial Alignment, TTA: Text-Temporal Alignment, ISA: Image-Spatial Alignment, ITA:Image-Temporal Alignment)} 
\centering
\resizebox{\textwidth}{!}{%

\begin{tabular}{|l|c|cc|ccc|cc|cc|c|}
\hline
\multirow{2}{*}{Method} & \multirow{2}{*}{resolution}  &\multicolumn{2}{c|}{VQ} &\multicolumn{3}{c|}{MQ} & \multicolumn{2}{c|}{TF} & \multicolumn{2}{c|}{IF}&\multirow{2}{*}{OL}\\ 
\cline{3-11} 
&  &PS & RO/S& MS &FC&AM&TSA&TTA&ISA&ITA & \\ \hline
I2VGen-XL~\cite{zhang2023i2vgenxl}& $32\times720\times1280$&\colornumber{98.79}& \colornumber{72.85}&\colornumber{87.63}& \colornumber{63.20}&\colornumber{-11.61}& \colornumber{26.24}&\colornumber{27.30}& \colornumber{97.13}&\colornumber{73.76}&\colornumber{85.22}\\
VideoCrafter1~\cite{chen2023videocrafter1} &$16\times 576\times 1024$ &\colornumber{73.74}& \colornumber{11.45}&\colornumber{80.61}&\colornumber{20.92}&\colornumber{-12.52}&\colornumber{-3.66}&\colornumber{-3.05}&\colornumber{92.82}&\colornumber{54.35}&\colornumber{53.59} \\
SVD~\cite{blattmann2023stable} & $25\times 576\times 1024$ &\colornumber{28.11}&\colornumber{4.41}&\colornumber{79.06}&\colornumber{12.59}&\colornumber{-41.43}&\colornumber{-0.44}&\colornumber{-4.39}&\colornumber{-14.79}&\colornumber{-1.76}&\colornumber{11.86} \\
Pika beta~\cite{pika} & $ 72\times 576 \times 1024$  & \colornumber{55.11}&\colornumber{2.44}&\colornumber{16.34}&\colornumber{9.62}&\colornumber{8.09}&\colornumber{3.76}&\colornumber{6.26}&\colornumber{0.92}&\colornumber{2.14}&\colornumber{27.33}\\
Gen2~\cite{gen2} & $ 96\times 1536 \times 2816$   & \colornumber{-34.86}&\colornumber{-2.19}&\colornumber{-3.72}&\colornumber{1.75}& \colornumber{-14.64}& \colornumber{-1.09}&\colornumber{4.04}&\colornumber{-14.54}&\colornumber{3.17}&\colornumber{-2.73}\\ \hline

\end{tabular}
}
\label{tab:human_eval}
\end{table}

\paragraph{Human Evaluation}\footnotemark
\footnotetext{Done in December 18th. The compared I2VGen is the version released in September.}
Due to the fact that automatic metrics are not able to fully reflect an individual's subjective perception of video quality, we further conduct human evaluations.  
Since many projects are close sourced, in this paper, we chose to compare with accessible works, including open-source works I2VGen-XL~\cite{zhang2023i2vgenxl}, VideoCrafter1~\cite{chen2023videocrafter1}, SVD~\cite{blattmann2023stable}, and closed-source works Pika beta~\cite{pika} and Gen2~\cite{gen2} that we can obtain the results from website or discord. All of these are recent works and represent the current best level in text/image-to-video generation.  
For a fair comparison, except for SVD and Pika beta  which only support image input, all other works were kept consistent in terms of text and image inputs~(The images are generated from text prompt by SDXL1.0 and refiner). The comparison results are shown in  Figure~\ref{fig1:prefernce} and Table~\ref{tab:human_eval}. Figure~\ref{fig1:prefernce} shows a comparison of \textbf{O}verall \textbf{L}ikeness between videos generated by our model~($\mathcal{F}_A+\mathcal{F}_{SR}$) and those produced by other methods. We find that the videos generated by our method outperform open-source Text/Image-to-video models and the closed-source method Pika beta, and are on par with the closed-source method Gen2.
Table~\ref{tab:human_eval} records the winning rates for other sub-metrics. The formula for calculating the winning rate from GSB is $(G-B) / (G+S+B)$. 
The \textred{number>0} indicates our method is better, and the \textgreen{number<0} indicates the other method is better. We found that the prominent advantage of our method lies in its \textbf{FC}, which is due to our adoption of an editing paradigm for low-freedom video generation, benefiting $\mathcal{F}_A$ in producing more stable videos. Additionally, our generated videos exhibit superior \textbf{PS} compared to videos of similar resolution (except for gen2 that generates videos of much larger resolution). This is because we employ \textbf{BGN}, ensuring consistency between training and inference by directly predicting high-resolution videos from low-resolution ones.
One significant drawback of our generated videos is the \textbf{AM}, due to our current lack of filtering for static videos in the training data. Addressing this will be part of our future work.

\subsection{Ablation Studies}
\noindent 
\begin{minipage}[c]{0.6\textwidth} 
\centering
\includegraphics[width=\textwidth,keepaspectratio]{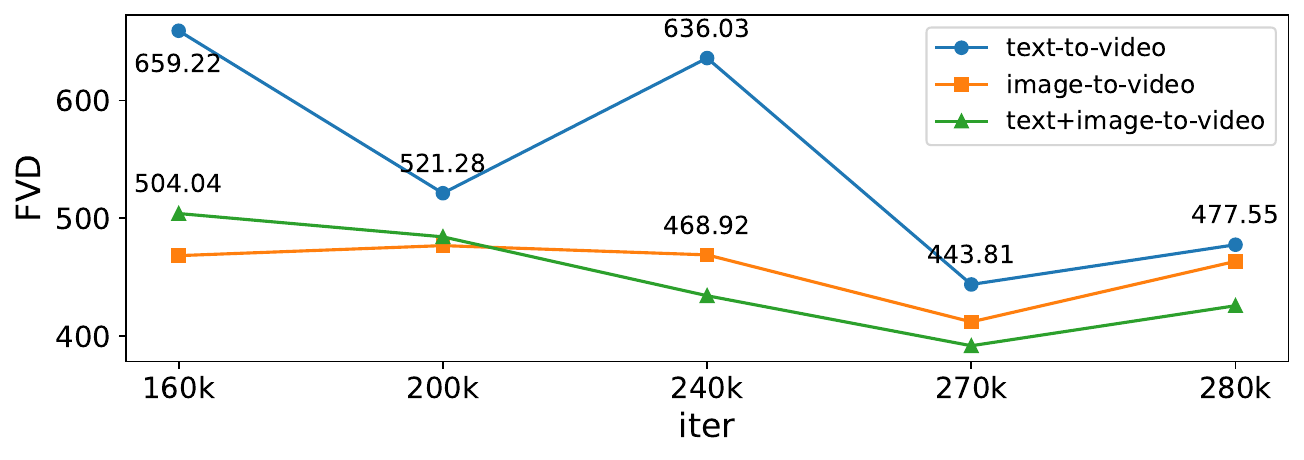}
 \captionof{figure}{FVD Scores on MSR-VTT during the Training Process of $\mathcal{F}_B$.}
\label{fig:ablation_base}

\end{minipage}%
\hspace{4mm} 
\begin{minipage}[c]{0.3\textwidth} 
\centering
\captionof{table}{FVD scores on MSR-VTT of $\mathcal{F}_A$ and  $\mathcal{F}_{SR}$ 
 that w/ or w/o BGN}
\resizebox{1.1\textwidth}{!}{%
  \begin{tabular}{c|c|c}
    \hline
    model&BGN&  FVD$\downarrow$  \\ \hline
    $\mathcal{F}_A$ & w/o BGN & 393.53\\
     $\mathcal{F}_A$ & w/ BGN & \textbf{369.27} \\ \hline
     $\mathcal{F}_{SR}$ & w/o BGN & 648.68\\
     $\mathcal{F}_{SR}$ & w/ BGN & \textbf{491.32} \\ \hline
  \end{tabular} 
  }\\
\label{tab:BGN}
\end{minipage}
\begin{figure*}[t]
  \centering
  \includegraphics[width=\linewidth]{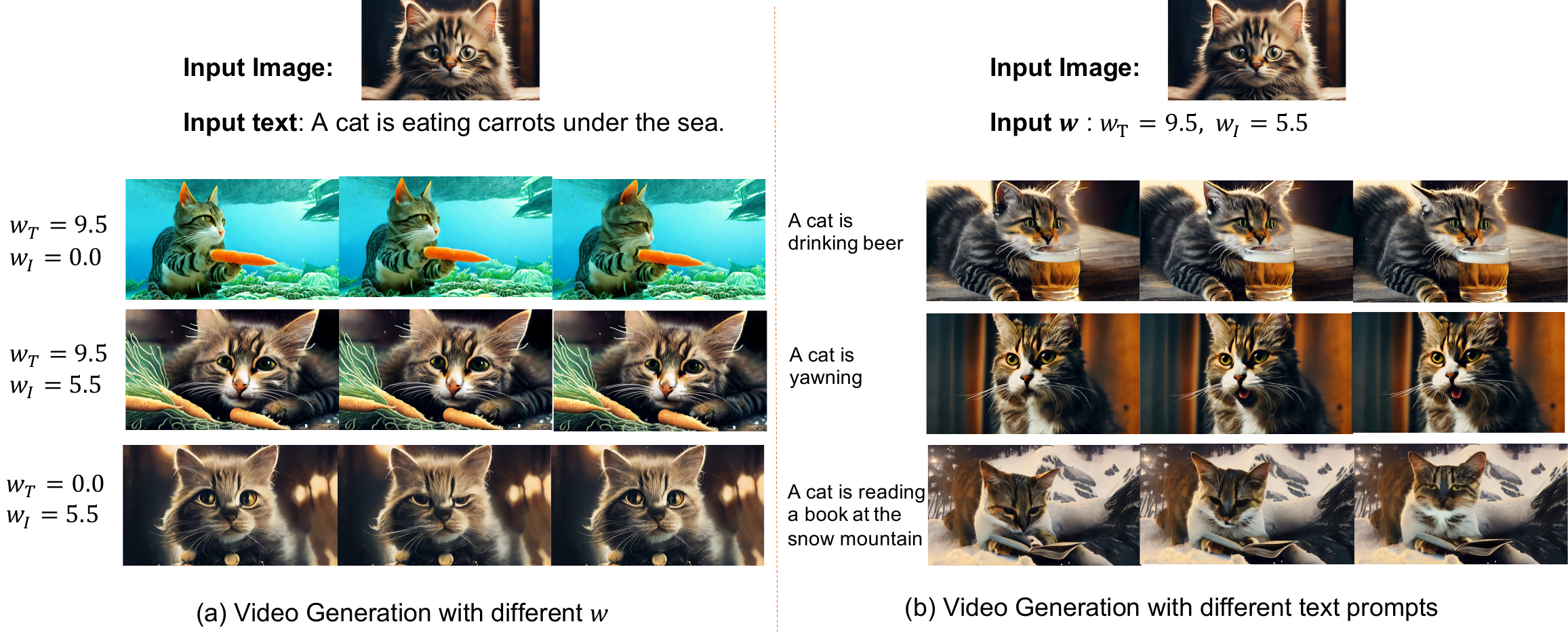}
  \caption{The generation cases of $\mathcal{F}_B$ with different classifier free guidance scale of text $w_T$ and $w_I$ and different text prompts.}
  \label{fig:ablation_conflict_condition}
\end{figure*}

\paragraph{Training Process of Base Model}
As our base model $\mathcal{F}_B$ emphasizes the conditional video generation with arbitrary combinations of text and images, a core question is whether the base model can maintain capabilities in text-to-video, image-to-video, and text/image-to-video generation simultaneously. Therefore, we take the checkpoints from the training process of $\mathcal{F}_B$ and test their performance in text-to-video, image-to-video, and text\&image-to-video generation with FVD. The results are shown in Figure \ref{fig:ablation_base}, where the overall trends of three curves are downward, indicating that the training process enhances the base model's ability to generate videos from text or images. This proves that for high-freedom video generation, multi-condition video generation can be integrated into one single model.

\begin{figure*}[t]
  \centering
  \includegraphics[width=\linewidth]{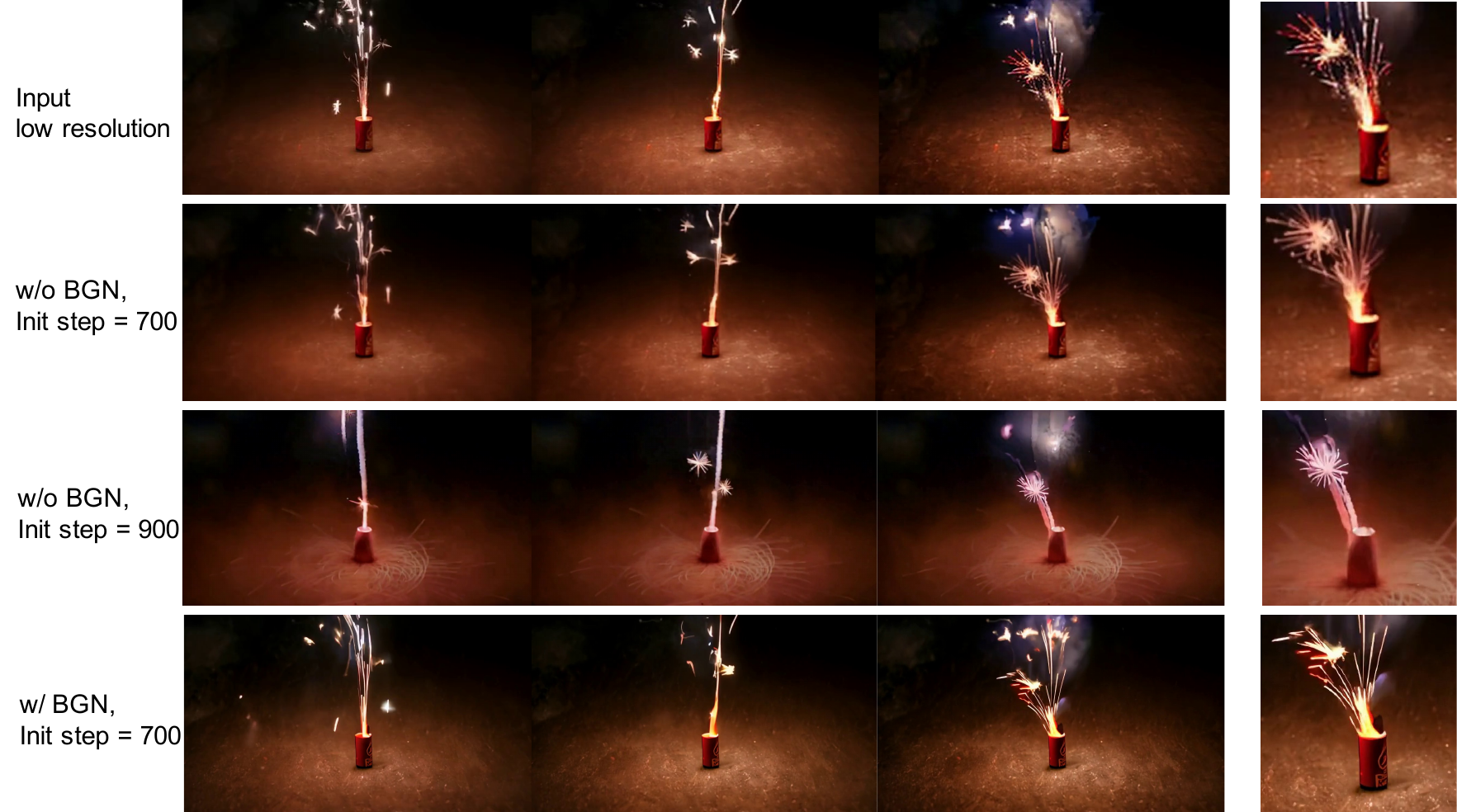}
  \caption{The generation cases of $\mathcal{F}_{SR}$ w/o or w/ BGN.}
  \label{fig:ablation_bn}
\end{figure*}

\paragraph{Biased Gaussian Noise}
To demonstrate that Biased Gaussian Noise (BGN) better suits low-freedom video generation tasks, we conducted ablation studies on the Animation Model $\mathcal{F}_A$ and the Video Super Resolution model $\mathcal{F}_{SR}$. The results, shown in Table~\ref{tab:BGN}, indicate that BGN enhances video quality in both Image Animation and Super Resolution, as evidenced by lower FVDs. It proves more beneficial for Super Resolution, a task with less freedom than Image Animation.
Figure~\ref{fig:ablation_bn} visualizes $\mathcal{F}_{SR}$'s performance with and without BGN. The first row shows the original, low-resolution input video. Rows 2 and 3 depict the outputs from $\mathcal{F}_{SR}$ without BGN, processed from the upscaled low-resolution input and subjected to $700$ and $900$ denoising steps, respectively. The fourth row presents the output from $\mathcal{F}_{SR}$ using BGN at timestep $t_m=700$ to $t_n=0$, illustrating how a low-resolution video upscaled to high-resolution can be denoised effectively after $700$ steps. Each row's far right offers a magnified view to better showcase the detail in the model-generated content.
Our observations indicate that absent BGN, a smaller initial noise step count results in less clarity (second row), while a larger count produces a clear yet inconsistent output due to noise overpowering the original content (third row). With BGN, the model directly predicts high-resolution videos from low-resolution inputs, achieving clarity and preserving original features (fourth row).
We also acknowledge that BGN's application can extend to other low-freedom video generation tasks, such as frame interpolation and video editing, which we aim to explore in future work.
\paragraph{Text\&Image Conditions}
Since our system is capable of generating videos that align both image and text flexibly, we explore the videos generated under different inference weights for these two conditions. Given text prompt $T$ and image condition $I$, the inference formula we use is $V_{out}=\mathcal{F}_B(\varnothing) + w_{T}(\mathcal{F}_B(T)-\mathcal{F}_B(\varnothing)) + w_{I}(\mathcal{F}_B(I)-\mathcal{F}_B(\varnothing))$. We adjust the classifier free guidance scale of text $w_T$ and image $w_I$, the generating videos are shown in Figure~\ref{fig:ablation_conflict_condition}-(a), we find that adjusting the $w_T$ and $w_I$ can bias the generated video towards the text or image conditions. Figure~\ref{fig:ablation_conflict_condition}-a shows that in row 1, $w_I=0$, $\mathcal{F}_B $ generates a video that is almost unrelated to the input image, while in row 3, $w_T=0$, $\mathcal{F}_B$ produces a video that is almost unrelated to the text. By adjusting both $w_T$ and $w_I$ to appropriate values, the second row's generated video retains the characteristics of the input image and is also aligned with the textual semantics. Based on this feature, our $\mathcal{F}_B$ can achieve different video generation with the same input image combined with different text prompts, as shown in Figure~\ref{fig:ablation_conflict_condition}-(b).
We have also explored whether $\mathcal{F}_A$ possesses similar properties. However, due to the concatenated image features having much more stronger constraints than text, the generated videos mainly rely on image semantics. Nevertheless, inputting consistent text helps to enhance the dynamic effects of the generated videos.

\section{Conclusion}
In this paper, we propose the UniVG system, designed for multi-task conditional video generation that leverages both text and images. We propose a novel categorization of models within our system based on generative freedom, distinguishing between high-freedom and low-freedom video generation. The high-freedom component of UniVG features a base model capable of modulating the influence of text and images to produce videos under varying semantic conditions. For low-freedom video generation, UniVG includes an Image Animation model and a Super Resolution model, which generate videos closely pixel-aligned with the input. In low-freedom generation, we propose Biased Gaussian Noise to replace the standard random Gaussian noise,  facilitating a more direct connection between the conditional and the target distributions. Our experiments show that our system outperforms existing methods in objective assessments and matches Gen2 in subjective evaluations.

\bibliographystyle{unsrt}  
\bibliography{references}

\end{document}